\documentclass[runningheads]{llncs}

\usepackage{amssymb}
\usepackage{graphicx}

\usepackage{url}
\urldef{\mailsa}\path|adrian.paschke@gmx.de|
\newcommand{\keywords}[1]{\par\addvspace\baselineskip
\noindent\keywordname\enspace\ignorespaces#1}

\begin{document}

\mainmatter  % start of an individual contribution

\title{Typed Hybrid Description Logic Programs with Order-Sorted Semantic Web Type Systems based on OWL and RDFS}

\titlerunning{ContractLog: A DL-Typed Logic Programming Language}

\author{Adrian Paschke}
% if the list of authors exceeds the space for a headline,
% an abbreviated author list is needed
\authorrunning{Adrian Paschke}

% the affiliations are given next
\institute{Internet-based Information Systems, Dept. of Informatics, TU Munich, Germany\\
 \mailsa\\
 Technical Report December 2005}

%\toctitle{Typed Logic} \tocauthor{Adrian Paschke}
\maketitle{

\begin{abstract}
In the recent years rule-based programming in terms of declarative
logic programming has formed the basis for many Artificial
Intelligence (AI) applications and is well integrated in the
mainstream information technology capturing higher-level decision
logics. Typically, the standard rule systems and rule-based logic
programming languages such as Prolog derivatives are based on the
untyped theory of predicate calculus with untyped logical objects
(untyped terms), i.e. the logical reasoning algorithms apply pure
syntactical reasoning. From a rule engineering perspective this is
a serious restriction which lacks major Software Engineering
principles such as data abstraction or modularization, which
become more and more important when rule applications grow larger
and more complex. To support such principles in logic programming
and capture the rule engineer's intended meaning of a logic
program, types and typed objects play an important role. Moreover,
from a computational point of view, the use of types drastically
reduces the search space, i.e. proofs can be kept at a more
abstract level and it offers the option to restrict the
application of rules and to control the level of generality in
queries. In this paper I introduce a multi-sorted logic which
allows using external Semantic Web type systems for term typing in
rules. Using Semantic Web ontologies as type systems facilitates
exchange of domain-independent rules over domain boundaries via
dynamically typing and mapping of explicitly defined type
terminologies. That is, rule description can take advantage of the
DL representation and reasoning capabilities, where different type
systems, assigning different domain-specific vocabularies to
rules, can be merged or disjoined and used dynamically at runtime.
Until recently, the use of Semantic Web languages such as RDFS,
OWL Lite or OWL DL has been limited primarily to representing Web
contents and previous works towards integration of rules and
ontologies in the Semantic Web mainly focus on extending the
general expressiveness of either the rule language or the ontology
language. I elaborate on a specific application, namely DL-typed
unification which integrates polymorphic order-sorted type
inference into the semantics of hybrid description logic programs.
This approach provides a technical separation with minimal
interfaces (based on query entailment) between the inferences in
the logic programming component and the description logic
inferences, which results in more efficient computations, enhanced
language expressiveness, a flexible and robustly decidable hybrid
integration, even in the case where the rule language is far more
expressive than Datalog. The multi-sorted logic supports subtypes,
so called order-sorted type systems, where types are analyzed at
run-time possibly permitting ad-hoc polymorphism of typed
variables, i.e. variables might change their types during runtime.
I present syntax and semantic of my hybrid DL-typed logic
programming language. I have developed a prototype system as part
of the ContractLog KR which supports typed unification and hybrid
knowledge bases and integrate the backward-reasoning
ContractLog/Prova rule engine with the Semantic Web API Jena and
the OWL-DL reasoner Pellet. Finally, I illustrate the integration
of types into the Rule Markup Language RuleML with serialization
of RDFS/OWL based type systems in XML syntax resp. XML/RDF syntax.
In contrast to homogeneous integration approaches such as DLP or
hybrid approaches based on additional DL constraints
(DL-atoms/queries) my hybrid approach benefits from the
prescriptive typing approach and the typed unification using an
external tableaux DL-reasoner (Jena/Pellet) for DL type checking
as a black box from the view of a rule engineer.

\end{abstract}

\section{On the Need for Semantic Web Types in Declarative Logic Programming}

Traditional logic programming languages such as most Prolog
derivatives are typically purely based on the untyped theory of
predicate calculus with untyped logical objects (untyped terms).
That is, the logical reasoning algorithms apply pure syntactical
reasoning and use flat untyped logical objects (terms) in their
rule descriptions. From the engineering perspective of a
rule-based SLA specification this is a serious restriction which
lacks major Software Engineering (SE) principles such as
\textit{data abstraction} or \textit{modularization}. Such
principles become important when rule applications grow larger and
more complex, are engineered and maintained by different people
and are managed and interchanged in a distributed environment over
domain-boundaries. To support such SE principles in declarative
programming and capture the rule engineer's intended meaning of a
contract, types and typed objects play an important role. Types
are an appropriate way to integrate domain specific ontologies
into dynamic domain-independent rules. As a result, such typed
rules are much easier to interchange between domain-boundaries in
distributed environments such as the Semantic Web where the domain
vocabularies are expressed as webized ontologies written in
expressive ontology languages such as RDF(S) or OWL. From a
descriptive point of view types attach additional information,
which can be used as \textit{type constraints} for selecting
specific goals, i.e. they constrain the level of generality in
queries, lead to much smaller search spaces and improve the
execution efficiency of query answering.

In the following I describe a multi-sorted logic with a
order-sorted typed unification as operational semantics which
allows using external DL type systems, i.e. Semantic Web
ontologies resp. knowledge bases, for term typing in rules. In
contrast to existing hybrid integration approaches which use
additional DL constraints or atoms/queries in the body of rules I
propose a prescriptive typing approach where types are direct
properties of the logical formulas, i.e. are directly attached to
rule terms. The typed order-sorted unification, which supports
dynamic type checking of terms during unification, provides a
built-in technical separation between the inferences in the rule
component and the type checks in the DL type system which results
in:

\begin{itemize}
    \item in more efficient computations, since the type restrictions directly apply during the  unification of typed terms
    \item higher flexibility with minimal interfaces (based on entailment) between rules component and the type system
    \item robustly decidable combinations, even in case where the rule language is far more expressive than Datalog and no safeness restrictions apply
    \item enhanced language expressiveness, e.g. ad-hoc polymorphism  with overloading is possible, type casting during order-sorted unification, data abstraction and modularization
    \item existing implementations can be reused
    \item rules can be more intuitively modelled and easily combined and interchanged
\end{itemize}

The further paper is structured as follows: I first review the
history of types in logic programming languages and describe basic
concepts in section 2. Then, in section 3, I elaborate on the
integration of rules and ontologies. In section 4 I describe the
syntax of my DL-typed rule language and in section 4 and 5 its
declarative and operational semantics, respectively. In section 6
I will discussion implementation and integrating of types into
rule markup languages and will discuss the ContractLog approach
towards a hybrid DL-typed logic. Finally, I will conclude this
paper with a short summary in section 7.

\section{Types in Logic Programming}

\begin{definition} (\textbf{Type System}) A type system \cite{Cardelli1997} is responsible for
assigning types to variables and expressions. It uses the type
expressions for static type checking at compile time and/or
dynamic type checking at runtime. Type systems typically define a
type relation of the form $t:r$, denoting that the term $t$ is of
type $r$. The primary purpose of a type system is to prevent
typing errors which violate the intended semantics of a rule or
function in a LP.
\end{definition}

That is, a type system is used to ensure robustness of a LP. By
ensuring that the structure is well-defined types enable a more
disciplined and regular engineering process and facilitate
modularity and partial specification of the intended use of the
logical functions and their arguments in a logic program. It has
been demonstrated that types play an important role to capture the
programmer's intended meaning of a logic program, see e.g.
\cite{Naish1992} and that they can be used to dramatically reduce
the rule search space - see e.g. Schubert's Steamroller Problem
which used to illustrate this advantage \cite{Stickel1986}.

The theory of types in LP has been studied in several works and
many different approaches for the introduction of types into logic
programming have been proposed reaching from many-sorted or
order-sorted systems with sub-typing to ad-hoc and parametric
polymorphic typing and combinations such as parametric polymorphic
order-sorted type systems. Most of the works on type systems and
their properties are based on the theory of $\lambda$-calculus
\cite{Barendregt1988} which gives some fundamentals for reasoning
about types in functional languages and which has been generalized
to different programming languages such as object-oriented or
declarative logic programming languages. Based on this theory one
of the most well-known type systems is the Hindley-Milner type
system \cite{Milner1978}. Different forms of type declarations
have been proposed such as declarations which use a rather general
constraint language \cite{Hohfeld1988}, logical formulas
\cite{Xu1988,Naish1992}, regular sets
\cite{Mishra1984,Mishra1985,Heintze1992,Dart1992,Apt1993},
equational specifications \cite{Hanus1992} or typed terms over a
order-sorted structure
\cite{Mycroft1984,Lakshman1991,Smolka1989b,Hill1992,Hanus1991}
(see e.g. \cite{Pfenning1992} for an overview).

In general, the works can be classified into two different views
on types in logic programming, namely \textit{descriptive types}
and \textit{prescriptive} types \cite{Reddy1988} (a.k.a. explicit
and implicit types or syntactic and semantic typing). Early work
on types in logic programming mainly concentrate on descriptive
type systems, e.g.
\cite{Dart1992a,Mishra1984,Hentenryck1994,Zobel1987} which attach
type information with programs without changing the language used
and without affecting the meaning of logic programs. These
descriptive approaches are seeking to approximate the structure of
a program for use by an optimizing compiler at compile time. For
instance, Mycroft and O'Keefe \cite{Mycroft1984} demonstrated that
the polymorphic type discipline of Milner \cite{Milner1978} can be
represented in pure Prolog, where the type declarations for
variables occur outside of the clauses and do not change the
semantics of the pure Prolog program. Dietrich and Hagl's
\cite{Dietrich1988} extend this approach with input/output mode
declarations and subsorts. In contrast, prescriptive type systems,
e.g. Gödel \cite{Hill1994}, Typed Prolog \cite{Lakshman1991},
$\lambda$-Prolog \cite{Miller1986,Miller1991}, consider types as
properties of the formulas one wants to give a meaning to, i.e.
they use a \emph{typed logic} for programming leading to languages
with higher expressiveness. The purpose is to identify ill-typed
programs, so that the actual semantics of a program satisfies the
intended semantics of the rule engineer. Lakshman and Reddy have
redefined the Mycroft-O'Keefe type discipline as Typed Prolog
\cite{Lakshman1991} which adopts the prescriptive view and gives a
semantics to the typed logic programs. Several other works follow
this approach of defining semantics for prescriptive typed LPs,
e.g. \cite{Heintze1992,Hill1992,Lakshman1991}.

These semantics approaches, to which also this ContractLog work
contributes, base logic programming with types on typed logics
that use sorts (type definitions). Following the terminology of
abstract data types \cite{Goguen1978,Ehrig1985} in
\textit{many-sorted type systems} sorts are defined by their
constructors, i.e.  the constituent elements of a sort. The sorts
are used to define the number of arguments and their types for
every predicate in a logic program. In the many-sorted case sorts
are not allowed to have subsort relations between them and
accordingly type checking can be done statically at compile time,
e.g. realized by a preprocessor without any extensions to the
underlying unsorted predicate logic. In the \textit{order-sorted
approach} subsort hierarchies are supported typically by the use
of a \emph{order-sorted unification} (typed unification) in order
to incorporate some form of \emph{subtyping polymorphism} for
untyped variables which assume the type of a unified typed term at
runtime. A first order-sorted logic was given by Oberschelp
\cite{Oberschelp1962} and an order-sorted algebra was developed by
Goguen et. al. \cite{Goguen1987,Smolka1989} which forms the basis
for the language Eqlog \cite{Goguen1986}. An extended order-sorted
algebra with error-handling was proposed by Gogolla
\cite{Gogolla1986}. Several other order-sorted approaches have
been described using order-sorted unification
\cite{Walther1987,Huber1987}. Different forms of polymorphism such
as \textit{generic polymorphism} (see e.g. ML programming language
\cite{Milner1978}), \textit{ad-hoc polymorphism} or
\textit{parametric polymorphism} haven been introduced into logic
programming. For a discussion of the differences between
parametric and ad-hoc polymorphism see e.g. \cite{Strachey2000}.
In the context of polymorphism terms (variables) are authorized to
change their types dynamically at runtime, which makes static
compile-time analysis insufficient in general. If the type system
permits ad-hoc polymorphism, the unifiers for terms are determined
by their types and the procedures (logical functions) being
invoked are dependent on this information, i.e. the types affect
the question of unifiability in an intrinsic way and the
computation process must use some form of typed unification
procedure to ensure type correctness also during runtime. The
types are needed to determine the existence of unifiers and hence
also the applicability of clauses. An interesting aspect of this
typed semantics is that it enables overloading of clauses leading
to different functional variants which are selected dynamically at
runtime according to the types of the queries. As a result, the
specific inferences that are performed and the correct answers to
queries are directly related to the types of the terms in the
program clauses and the answers to queries not only display the
bindings of variables, but also their types. Typed unification has
been studied for order-sorted and polymorphic type systems, see
e.g. Typed Prolog \cite{Lakshman1991}, Protos-L
\cite{Beierle1989}, $\lambda$-Prolog extensions \cite{Kwon1993}.
\emph{Order-sorted unification} extends the usual term unification
with additional dynamic type checking. In a nutshell, the basic
idea of sorted unification of two typed variables is to find the
greatest lower bound (glb) of their types based on the type
hierarchy with subtype relationships, failing if it does not
exist. In other words, the unification algorithm tries to find the
glb of two sort restrictions yielding a variable whose sort
restriction is the greatest common subsort of the two sorts of the
unified terms in the given sort hierarchy. This typing approach
provides higher levels of abstractions and allows \emph{ad-hoc
polymorphism} wrt \emph{coercion}, i.e. automatic type conversion
between subtypes, and \emph{overloading}, i.e. defining multiple
functions (rules with the same head but different types) taking
different types, where the unification algorithm automatically
does the type conversion and calls the right function (unifies a
(sub-)goal with the right rule head). \emph{Parametric polymorphic
types} allow to parameterize a structured sort over some other
sort, i.e. types and predicates can be written generically so that
they can deal equally with any object without depending on their
types, in contrast to the many-sorted case, where for each
predicate variant having a different type a sort definition must
be given explicitly. Typically, parametric polymorphism still
maintains full static type-safety on a syntactic level, however
there are approaches with a semantic notion of polymorphic types,
e.g. order-sorted parametric polymorphic typed logics which have
to take the type information into account at runtime and hence
require an extended unification algorithm with type inferencing,
as in the order sorted case. These polymorphic type systems being
very complicated artifacts both theoretically and computational
wise and have been primarily designed for use in the context of
higher-order logic programming. The emphasis in these higher-order
type languages has been on describing conditions under which the
computationally expensive type analysis can be avoided at runtime
which often amounts to banishing ad hoc polymorphism and applying
several restrictions to function symbols which must be type
preserving. Recent works on types for LPs have concentrated on
implementation techniques for efficiently checking or inferring
types at runtime, in particular polymorphic types, e.g. by means
of abstract interpretations \cite{Lu1998} or constraint solving
techniques \cite{Demoen1999}.

\section{Description Logic Type System}

Until recently, the use of Semantic Web ontology languages such as
OWL \cite{McGuinness2004} or RDFS \cite{Brickley2004} has been
limited primarily to define meta data vocabularies and add
semantic machine-readable meta data to Web pages enabling
automated processing of Web contents. Both, Description Logics
\cite{Nardi2002} which form the logical basis of Semantic Web
serialization languages such as OWL and (Horn) LPs (restricted to
function-free Datalog LPs) are decidable fragments of first order
logic, however for the most part with orthogonal expressive power.
Whereas OWL is restricted to unary resp. binary axioms, but e.g.
provides classical/strong negation under open world assumption
(OWA) and existentially as well as universally quantified
variables, Datalog LPs allow n-ary axioms and (non-monotonic)
negation, but are restricted to universal quantifications and are
therefore not able to reason about unknown individuals. Clearly,
both approaches can benefit from a combination and several
integration approaches have been proposed recently.

The works on combining rules and ontologies can be basically
classified into two basic approaches: \textit{homogeneous and
heterogeneous integrations}. Starting from the early Krypthon
language \cite{Brachman1985} among the heterogeneous approaches,
which hybridly use DL reasoning techniques and tools in
combination with rule languages and rule engines are e.g. CARIN
\cite{Levy1996}, Life \cite{Ait-Kaci1991}, Al-log
\cite{Donini1991}, non-monotonic dl-programs \cite{Eiter2004} and
r-hybrid KBs \cite{Riccardo2005}. Among the homogeneous approaches
which combine the rule component and the DL component in  one
homogeneous framework sharing the combined language symbols are
e.g. DLP \cite{Grosof2003}, KAON2 \cite{Motik2005} or SWRL
\cite{Horrocks2004}. Both integration approaches have pros and
cons and different integration strategies such as reductions or
fixpoint iterations are applied with different restrictions to
ensure decidability. These restrictions reach from only allowing
the intersection of DLs and Horn rules \cite{Grosof2003}, to
leaving full syntactic freedom for the DL component, but
restricting the rules to \textit{DL-safe rules} \cite{Motik2005},
where DL variables must also occur in a non DL-atom in the rule
body, or \textit{role-safe rules} \cite{Levy1996}, where at least
one variable in a binary DL-query in the body of a hybrid rule
must also appear in a non-DL atom in the body of the rule which
never appears in the consequent of any rule in the program or to
tree-shaped rules \cite{Heymans2005}. Furthermore, they can be
distinguished according to their information flow which might be
\textit{uni-directional} or \textit{bi-directional}. For instance,
in homogeneous approaches bi-directional information flows between
the rules and the ontology part are naturally supported and new DL
constructs introduced in the rule heads can be directly used in
the integrated ontology inferences, e.g. with the restriction that
the variables also appear in the rule body (safeness condition).
However, in these approaches the DL reasoning is typically solved
completely by the rule engine and benefits of existing tableau
based algorithms in DL reasoners are lost. On the other side,
heterogenous approaches, benefit from the hybrid use of both
reasoning concepts exploiting the advantages of both (using LP
reasoning and tableaux based DL reasoning), but bi-directional
information flow and fresh DL constructs in rule heads are much
more difficult to implement. In summary, the question wether the
Semantic Web should adopt an homogeneous or heterogenous view is
still very much at the beginning and needs more investigation.

In ContractLog I have implemented support for both homogeneous and
heterogeneous integration approaches, in the so called
\textit{OWL2Prova API}, which can be configured and used in rule
representation. Nevertheless, in the context of term typing using
ontologies or object class hierarchies as order-sorted type
systems, I use a hybrid prescriptive typing approach which
exploits highly optimized external DL reasoner for type checking
by means of subsumption checking and instance inference permitting
also equivalence reasoning or inner anonymous existentials. In
contrast to homogenous approaches such as SWRL (the union of
OWL-DL and Datalog \cite{Horrocks2004}) or DLP (the intersection
of restricted OWL and Datalog \cite{Grosof2003}) the heterogeneous
approach has the advantage that is does not need any non-standard
homogenous reasoning services, but may reuse existing
implementations exploiting e.g. highly optimized external DL
reasoner for DL reasoning tasks and a rule engine for the rule
reasoning tasks. In most hybrid approaches, which combine
ontologies and rules, the interaction between the subsystems is
solved by additional \emph{constraints} in the body of (Datalog)
rules, which restrict the values of variables and constants in the
\emph{constraint clauses} to range over the instances of the
externally specified DL concepts. Although this constraint
solution to some extend ensures expressiveness in the sense that
it allows to reuse the binary relations (classes and object
properties) defined in the external ontology, it has some serious
drawbacks in the context of term typing, since this approach
leaves the usual operational semantics of resolution and
unification unchanged. Hence, the constraints apply only in the
body of a rule according to the selection function of the LP
inference algorithm which typically selects the "left-most"
unified literal as next subgoal. As a result, according to the
standard depth-first strategy the type constraints are used
relatively late in the derivation process. In particular, they do
not apply directly during the unification between (sub)goals and
rule heads, which leads to large and needless search spaces
(failed search trees) and might lead to unintended results since
there is no way to directly verify and exclude the unification of
typed free (sub)goals and typed rule heads which do not match
according to their type definitions. Moreover, fresh typed
constant terms, i.e. new DL instances of a particular DL type
(class), can not be introduced directly in rule heads since this
would require special semantics and formalizations, e.g., with
conjunctive rule heads consisting of a literal which defines the
individual and another literal which defines its type (or
Lloyd-Topor transformations). For instance, a rule with a variable
$X$ of type $C$ in the rule head, would need an extra constraint,
e.g. $type(X,C)$ in the body, e.g.: $p(X) \leftarrow q(X),
type(X,C)$. Obviously, such a constraint rule is dependent on the
order of the body atoms: if $type(X,C)$ is before $q(X)$ the
variable $X$ might not be bound to a ground individual and hence
can not be verified; on the other hand, if it is after $q(X)$ it
does not directly constrain the subgoal $q(X)$, which might be
possibly very deep including many variable bindings which are not
of type $C$.

In contrast to these constraint approaches or approaches which
apply explicit additional DL-queries/atoms in the rules' body for
querying the type system, I have implemented a hybrid typed logic
in ContractLog where dynamic type checking for prescriptively
typed term is directly integrated into the unification process,
i.e. directly implemented within the operational semantics, rather
than indirectly through the resolution-based inference mechanism
solving the additional constraint/query terms in the body of
rules. The interaction during typed unification between the rule
component and the DL type system is based on entailment. As a
result, the hybrid typed method provides a built-in technical
separation between the inferences in the rule component and the
type checks in the DL type system (resp. Java type system) which
results in:

\begin{itemize}
    \item in more efficient computations, since the type restrictions directly apply during the  unification of typed terms
    \item higher flexibility with minimal interfaces (based on entailment) between rules component and the type system
    \item robustly decidable combinations, even in case where the rule language is far more expressive than Datalog and no safeness restrictions apply
    \item enhanced language expressiveness, e.g. ad-hoc polymorphism  with overloading is possible, type casting during order-sorted unification, data abstraction and modularization
    \item existing implementations can be reused
    \item rules can be more intuitively modelled and easily combined and interchanged
\end{itemize}

In the following two sections I will describe the syntax and
semantics of the typed logic which I have implemented in the
ContractLog KR. In particular I will elaborate on hybrid
description logic programs (\textit{hybrid DLPs}), namely
description logic Semantic Web type systems (DL type systems with
DL-types) which are used for \emph{term typing of LP rules based
on a polymorphic, order-sorted, hybrid DL-typed unification as
operational semantics for hybrid DLPs}.

\subsection{Syntax of Typed ContractLog}

A ContractLog LP is an extended LP (ELP) \cite{Lifschitz1992},
i.e. a LP with monotonic explicit negation and non-monotonic
default negation. An ELP is a set of clauses of the from $H
\leftarrow B$, where $H$ is a literal over $L$ called the head of
the rule, and $B$ is a set of literals over $L$ called the body of
the rule. A literal $B_{i}$, which might be default negated "$\sim
B_{i}$", is either an atom or the negation $\neg$ of an atom,
where $\sim $ denotes default negation written as $not(...)$ and
$\neg$ is denoted as explicit negation written as $neg(...)$.
Roughly, default negation means, everything that can not be proven
as true is assumed to be false. A rule is called a fact if it only
consists of the rule head $H$. An atom is a n-ary formula
containing terms $p(a, X, f(...))$, where $p$ is the predicate
name. A term is either a constant $a$, a variable $X$ or a n-ary
complex term/function $f(...)$. A goal/query $G$ is a headless
clause defining a conjunction of literals (positive or negative
atoms) $L_{1} \wedge .. \wedge L_{i}$ where each $L_{i}$ is called
a subgoal. A \textit{query} is embedded in the built-in function
$:-sovle(...)$ or $:-eval(...)$.

ContractLog assumes not just a single universe of discourse, but
several domains, so called sorts (types). I first describe the
basic extension of the language towards a multi-sorted logic, i.e.
the extension of the signature and the variables of the alphabet
with sorts.

\begin{definition} (\textbf{Multi-sorted Signature}) A multi-sorted
signature $S$ is defined as a tuple $\langle S_{1},..,S_{n},
\overline{P}$, $\overline{F}, arity, \overline{c}, sort \rangle$
where:

\begin{enumerate}
    \item  $S_{1},..,S_{n}$ is a list of type symbols called sorts (types)
    \item $\overline{P}$ is a finite sequence of predicate symbols
    $\langle P_{1},..,P_{n} \rangle$.
    \item $\overline{F}$ is a finite sequence of function symbols $\langle F_{1},..,F_{m} \rangle$
    \item For each $P_{i}$ respectively each $F_{j}$, $arity(P_{i})$ resp. $arity(F_{j})$ is a non-zero natural number
    denoting the arity of $P_{i}$ resp. $F_{i}$.
    \item $\overline{c} = \langle c1,..,c_{o} \rangle$ is a finite
    or infinite sequence of constant symbols.
    \item $sort$ is a function that associates with each predicate, function or constant its sorts
\end{enumerate}

The function $sort$ is defined as follows:

\begin{itemize}
    \item if $p$ is a predicate of arity $k$, then $sort(p)$ is a k-tuple of sorts $sort(p)=(X_{1},..X_{k})$ where each $X_{i}$ is of some sort $S_{j}$.
    \item if $f$ is a function of arity $k$, then $sort(f)$ is a $k+1$-tuple of sorts defining the sorts of the domain and the range of $f$
    \item if $c$ is a constant, then $sort(c)$ gives the sort of $c$.
\end{itemize}

I define the following three basic types of sorts

\begin{enumerate}
    \item primitive sorts are given as a fixed set of primitive data types such as integer,
    string, etc.
    \item function sorts are complex sorts constructed from
    primitive sorts (or other complex sorts) $S_{1} \times ...
    \times S_{n} \rightarrow S_{n+1}$
    \item Boolean sorts are (predicate) statement of the form $S_{1} \times ...
    \times S_{n}$
\end{enumerate}

\end{definition}

Additionally, each variable $V_{j}$ in the language with a
multi-sorted signature is associated with a specific sort:
$sort(V_{j})=S_{i}$. The intuitive meaning is that a predicate or
function holds only if each of its terms is of the respective sort
given by $sort$. A binary equality predicate $=$ exists in the
language. I write $t_{1}=t{2}$ instead of $=(t1,t2)$.

\begin{definition} (\textbf{Multi-sorted Logic}) A multi-sorted logic associates which each term, predicate and function a particular
sort:

\begin{enumerate}
    \item Any constant or variable $t$ is a term and its sort is given by $sort(t)$
    \item Let $F(t_{1},..,t_{n})$ be a function then it is a term of sort $S_{n+1}$ if $sort(F)= \langle S_{1},..,S_{n},S_{n+1} \rangle$, i.e. $F$ takes argument of sort $S_{1},..,S_{n}$ and returns arguments in sort $S_{n+1}$.
\end{enumerate}

\end{definition}

I now extend the language to consider external sort/type alphabets
and combined signatures as basis for combined knowledge bases and
the integration of external type systems into rule bases.

\begin{definition} (\textbf{Type alphabet}) A type alphabet $T$ is a finite set of monomorphic sort/type symbols built over the distinct set of terminological class concepts
of a (external type) language $\Sigma$.
\end{definition}

Informally, a typed ContractLog LP consists of an extended LP with
typed terms and a set of external (order-sorted) type systems in
which the types (sorts) are defined based on their type alphabets.
An external type system might possibly define a complete knowledge
base with types/sorts (classes) and individuals associated with
these types (instances of the classes). The combined signature is
then the union of the two (or more) signatures, i.e. the
combination of the signature of the rule component and the
signatures of the external type systems / knowledge bases
including their type alphabets, their functions and predicates and
their individuals.

\begin{definition} (\textbf{Combined Signature}) A combined signature $\Sigma$ is the union of all its
constituent finite signatures: $\Sigma=\langle S_{1} \cup .. \cup
S_{n} \rangle$
\end{definition}

Based on this definition of a combined signature I now describe
the concrete syntax of typed ContractLog. Using equalities
ContractLog implements a notion of default inequality for the
combined set of individuals/constants which leads to a less
restrictive unique name assumption:

\begin{definition} (\textbf{Default Unique Name Assumption}) Two ground terms are assumed to be unequal, unless equality
between the terms can be derived.
\end{definition}

The type systems considered in ContractLog are order-sorted.

\begin{definition} (\textbf{Order-sorted Type System}) A finite order-sorted type system $T$ comes with a partial order $\leq$, i.e. $T$ under $\leq$ has a
greatest lower bound $glb(r_{1},r_{2})$ for any two types $r_{1}$
and $r_{2}$ having a lower bound at all. Since $T$ is finite also
a least upper bound $lub(r_{1},r_{2})$ exists for any two types
$r_{1}$ and $r_{2}$ having a upper bound at all.
\end{definition}

Currently, ContractLog supports two external order-sorted type
systems (with sub-type relations), one is \textit{Java} and the
other one are \textit{Semantic Web ontologies} (defined in OWL or
RDFS) respectively Description Logic KBs. In the following I
describe the syntax of typed ContractLog for external DL-type
systems. I have chosen a \textbf{prescriptive typing approach},
where types are direct properties of the logical formulas, i.e.
are directly attached to terms in a type relation $t:r$, denoting
that a term $t$ has a type $r$. This has several advantages for
the semantics.

ContractLog supports webized Semantic Web ontologies defined in
RDFS or OWL (OWL Lite or OWL DL) as type systems. That is, the
combined signature consisting of the finite signature of the rule
component and the finite signature of the ontology language(s).
Note that the approach is general and can be extended to different
DL languages defining DL KBs as type systems. Hence, in the
following I use the term DL also for Semantic Web ontology
languages and knowledge bases. For the reason of
understandability, I assume only one external type system in the
following.

The type alphabet $T$ is a finite set of monomorphic type symbols
built over the distinct set of terminological atomic concepts $C$
in a Semantic Web ontology language $L$, i.e. defined by the
atomic classes in the T-Box model. Note, that restricting types to
atomic concepts is not a real restriction, because for any complex
concept such as ($C \sqcap D$) or ($C \sqcup D$) one may introduce
an atomic concept $A_{C}$, i.e. add the axiom $C \sqsubseteq
A_{C}$ to the T-Box, and use $A_{C}$ as atomic type instead of the
complex type. This approach is also reasonable from a practical
point of view since dynamic type checking must be computationally
efficient in order to be usable in a order-sorted typed logic with
possible very large rule derivation trees and typed unification,
i.e. fast type checks are crucial during typed term unification. I
assume that the type alphabet is fixed (but arbitrary), i.e. no
new terminological concepts can be introduced in the T-Box by the
rules at runtime. This ensure that I can also apply static type
checking on the used DL-types in ContractLog at compile time
(during parsing the LP script).

I use a prescriptive typing approach for the "has-type" relation
$t:r$. The set of constants/individuals is built over the set of
individual names in $L$, but I do not fix the constant names and
allow arbitrary fresh constants (individuals) (under default UNA)
to be introduced in the head of rules and facts. The precise
syntax is a follows:

\begin{definition} (\textbf{DL-typed Terms}) A type is a terminological concept/class defined in the type
system (T-Box model) possibly prefixed by a URI namespace
abbreviation separated by $\_$, i.e. $namespace\_Type$. A typed
term is denoted by the relation $t:r$ for typed variable terms and
$r:t$ for typed constant terms, i.e. $ns\_a:ns\_C$ denoting that
individual $ns\_a$ is of type $ns\_C$ or $X:ns\_C$, i.e. variable
$X$ is of type $ns\_C$.
\end{definition}

The type ontologies are typically provided on the Web under a
certain URL and types and individuals are represented as resources
having an webized URI. Defined namespace can be used to avoid name
conflicts and namespace abbreviations facilitate are more readable
language. Note, that fresh individuals which are introduced in
rules or facts apply locally within the scope of the
predicate/function and rule in which they are defined, i.e. within
a reasoning chain; in contrast to the individuals defined in the
A-box model of the type system which apply globally as individuals
of a class (type).

\begin{example}
\end{example}
\begin{verbatim}
% Import external type systems

import("http//../dl_typing/businessVocabulary1.owl").
import("http//../dl_typing/businessVocabulary2.owl").
import("http//../dl_typing/mathVocabulary.owl").
import("http//../dl_typing/currencyVocabulary.owl").

reasoner("dl"). % configure reasoner (OWL-DL=Pellet)


% Rule-based Discount Policy

discount(X:businessVoc1_Customer, math_Percentage:10) :-
    gold(X: businessVoc1_Customer).
discount(X: businessVoc1_Customer, math_Percentage:5) :-
    silver(X: businessVoc1_Customer).
discount(X: businessVoc1_Customer, math_Percentage:2) :-
  bronze(X: businessVoc1_Customer).

% Note that this rules use a different vocabulary
% if the types "Client" and "Customer" are equal
% both typed rule sets unify
% Class equivalence between both "types"
% is defined in the second OWL-DL ontology

gold(X:businessVoc2_Client) :-
   spending(X:businessVoc2_Client, S:currency_Dollar),
   S:currency_Dollar > currency_Dollar:1000.
silver(X:businessVoc2_Client) :-
   spending(X:businessVoc2_Client, S:currency_Dollar),
   S:currency_Dollar > currency_Dollar:500.
bronze(X:businessVoc2_Client) :-
   spending(X:businessVoc2_Client, S:currency_Dollar),
   S:currency_Dollar > currency_Dollar:100.


% Facts

spending(businessVoc1_Customer:Adrian, currency_Dollar:1000).
spending(businessVoc1_Customer:Aira, currency_Dollar:200).

% Query

:-solve(discount(X:businessVoc_Client, Y:math_Percentage)).
\end{verbatim}

The examples shows term typing with "plug-able" (imports of)
Semantic Web type vocabularies. Remarkably, $businessVocabulary1$
defines a type $Customer$ which is defined to be equal to the type
$Client$ in $businessVocabulary2$. Hence, both types unify and the
first three rules relate to the second three rules of the discount
policy as well as to the defined facts. The user can use both
types interchangeable to define queries on the hybrid KB, i.e.,
the domain-specific vocabulary of choice can be used.

Free DL-typed variables are allowed in facts. As I will describe
in the semantics section they act as instance queries on the
ontology, i.e. they query all individuals of the given type and
bind them to the typed variable. In addition ContractLog provides
a special query predicate which can be used in the body of rules
to interact with the ontology component and explicitly express
queries, such as concept membership, role membership or concept
inclusion on the DL knowledge base. The special query predicate
$rdf$ (implemented in owl.prova library) is used to query external
ontologies written in RDF(S) or OWL (OWL Lite or OWL DL).

\begin{example}
\end{example}
\begin{verbatim}
% Bind all individuals of type "Wine" to the variable "Subject"
%using the owl ontology WineProjectOWL.owl and the "rdfs" reasoner
rdf(
    "./examples/function_tests/owl/testdata/WineProjectOWL.owl",
    "rdfs",
    Subject,"rdf_type","http://www.owl-ontologies.com/unnamed.owl#Wine")

% Use the transitive reasoner and namespace abbreviations
rdf(
    "./rules/function_tests/owl/testdata/WineProjectOWL.owl",
    "transitive",
    Subject,"rdfs_subClassOf","default_Wine")
\end{verbatim}

The first argument specifies the URL of the external ontology. The
second argument specifies the external reasoner which is used to
infer the ontology model and answer the query. Note, that this
hybrid method using an external reasoner to answer a queries
provides a technical separation between the inferences in the
Description Logic part which is solved by an optimized external DL
reasoner and the Logic Programming components which is solved by
the rule engine. As a result the combined approach is robustly
decidable, even in case where the rule language is far more
expressive than Datalog. Moreover, the triple-based query language
also supports queries to plain RDF data sources, e.g. Dublin Core
meta data. The following predefined reasoner are supported:

\begin{itemize}
    \item "" | "empty" | null = no reasoner
    \item default = OWL reasoner
    \item transitive = transitive reasoner
    \item rdfs = RDFS rule reasoner
    \item owl = OWL reasoner
    \item daml = DAML reasoner
    \item dl = OWL-DL reasoner
    \item swrl = SWRL reasoner
    \item rdfs\_full = rdfs full reasoner
    \item rdfs\_simple = rdfs simple reasoner
    \item owl\_mini = owl mini reasoner
    \item owl\_micro = owl micro reasoner
\end{itemize}

User-defined reasoners can be easily configured and used. By
default the specified reasoners are used to query the external
models on the fly, i.e. to dynamically answer the queries using
the external reasoner. But, a pre-processing mode is also
supported. Here the reasoners are used to pre-infer the ontology
model, i.e. build an inferred RDF triple model where the logical
DL entailments such as transitive subclasses are already resolved
at compilation time. Queries then operate on the inferred model
and are hence much fast to answer, however with the drawback that
updates of the ontology model require a complete recompilation of
the inferred model.

\section{Semantics of Typed ContractLog}

\subsection{Declarative Semantics: Multi-Sorted Logic}

The semantics of a LP $P$ is typically defined wrt to the closed
Herbrand universe $B(P)$. I now extend the domain of discourse
towards a combined knowledge base defined over a combined
signature where individuals and types (sorts) from one or more
type systems outside of the fixed domain of the rule component are
taken into account. The semantics of the combined KB based on an
extended Herbrand Base is then defined wrt to the combined
signature.

\begin{definition} (\textbf{Combined Knowledge Base}) The combined
knowledge base of a type ContractLog LP $KB = \langle \Pi^{LP},
\Phi^{TS} \rangle$ consists of a finite set of (order-sorted) type
systems / type knowledge bases $\Phi^{TS} = \Phi^{TS}_{1} \cap ..
\cap \Phi^{TS}_{n}$ and a typed ContractLog KB $\Pi^{LP}$
represented as a defeasible extended typed LP.
\end{definition}

The combined signature of the combined KB is the union of all
constituent signatures, i.e. each interpretation of a ContractLog
LP has the set of ground terms of the combined signature as its
fixed universe.

\begin{definition} (\textbf{Extended Herbrand Base}) Let $KB = \langle \Pi^{LP},
\Phi^{TS} \rangle$ a typed combined ContractLog LP $P$. The
extended Herbrand Base of $P$, denoted $B(P)$, is the set of all
ground literals which can be formed by using the predicate symbols
in the combined signature with the ground typed terms in the
combined universe $U(P)$, which is the set of all ground typed
terms which can be formed out of the constants, type and function
symbols of the combined signature.
\end{definition}

The grounding of the combined KB is computed wrt the composite
signature.

\begin{definition} (\textbf{Grounding}) Let $P$ be a typed (combined) ContractLog
LP and $C$ its set of constant symbols in the combined signature.
The grounding $ground(P)$ consists of all ground instances of all
rules in $P$ w.r.t to the combined multi-sorted signature which can be obtained as follows:\\
The ground instantiation of a rule $r$ is the collection of all
formulae $r[X_{1}:S_{1}/t_{1},..,X_{n}:S_{n}/t_{n}]$ with
$X_{1},..,X_{n}$ denoting the variables and $S:{1},..,S_{n}$ the
types of the variables (which must not necessarily be disjoint)
which occur in $r$ and $t_{1},..,t_{n}$ ranging over all constants
in $C$ wrt to their types. For every explicit query/goal
$Q[Y_{1}:T_{1},..,Y_{m}:T_{m}]$ to the type system, being either a
fact with one or more free typed variables
$Y_{1}:T_{1},..,Y_{m}:T_{m}$ or a special query atom $rdf(...)$
with variables as arguments in the triple-like query, the
grounding $ground(Q)$ is an instantiation of all variables with
constants (individuals) in $C$ according to their types.
\end{definition}

The interpretation $I$ of a typed program $P$ then is a subset of
the extended Herbrand base $B(P)$.

\begin{definition} (\textbf{Multi-sorted Interpretation}) Let $KB= \langle \Pi^{LP},
\Phi^{TS} \rangle$ be a combined KB and $C$ its set of constant
symbols. An interpretation $I$ for a multi-sorted combined
signature $\Sigma$ consists of

\begin{enumerate}
    \item a universe $\mid I | = S_{1}^{I} \cup S_{2}^{I} \cup .. \cup S_{n}^{I}$, which is the union of the types (sorts), and
    \item the predicates, function symbols and constansts/individuals $C$ in the combined signature, which are interpreted in accordance with their types.
\end{enumerate}

The assignment function $\sigma$ from the set of variable $V$ into
the universe $| I |$ must respect the sorts/types of the variables
(in order-sorted type systems also subtypes). That is, if $X$ is a
variable of type $T$, then $\sigma(X) \in T^{I}$.  In general, if
$\phi$ is a typed atom or function in $\Pi^{LP}$ and $\sigma$ an
assignment to the interpretation $I$, then $I \models
\phi[\sigma]$, i.e. $\phi$ is true in $I$ when each variable $X$
of $\phi$ is substituted by the values $\sigma(X)$ wrt to its
type. Since the assignment to constant and function symbols is
fixed and the domain of discourse corresponds one-to-one with the
constants $C$ in the combined signature, it is possible to
identify an interpretation $I$ with a subset of the extended
Herbrand base.
\end{definition}

The assignment function is given as a query from the rule
component to the type system, so that there is a separation
between the inferences in a type system and the rule component.
Moreover, explicit queries to a type system defined in the body of
a  rule, e.g. ontology queries (special $rdf$ query or free
DL-typed facts) are based on this hybrid query mechanism. The
query interaction between the rules and the type system  is based
on entailment. I now define the notion of model for a typed
ContractLog LP

\begin{definition} (\textbf{Model}) et $KB= \langle \Pi^{LP},
\Phi^{TS} \rangle$ be a combined KB of a typed ContractLog LP $P$.
An interpretation $I$ is a model of a untyped ground atom $a \in
B(\Pi^{LP})$ or $I$ satisfies $a$, denoted $I \models a$ iff $a
\in I$. $I$ is a model for a typed ground atom $b \in
B(\Pi^{LP})$, or $I$ satisfies $b$, denoted $I \models b$, iff $b
\in I$ and for every typed term $t_{i}:s_{i}$ in $b$ the type
query $s_{i}(t_{i})$ (is $t_{i}$ of type $s_{i}$) is entailed in
$\Phi^{TS}$, i.e. $\Phi^{TS} \models s_{i}(t_{i})$ (in an order
sorted type system subtypes are considered, i.e. $t_{i}$ is of the
same or a subtype of $s_{i}$). $I$ is an interpretation of an
ground explicit query/goal $Q$ to the type system $\Phi^{TS}$ if
$\Phi^{TS} \models Q$.\\
$I$ is a model of a ground rule $r : H \leftarrow B$ iff $I
\models H(r)$ whenever $I \models B(r)$. $I$ is a model a typed
program $P$ (resp. a combined knowledge base $KB$), denoted by $I
\models P$, if $I \models r$ for all $r \in ground(P)$.
\end{definition}

The default semantics for ContractLog is extended well-founded
semantics. To support well-founded semantics for negated queries
to external type systems, i.e. the negated query $\sim Q$
succeeds, if the query fails, i.e. $\Phi^{TS} \nvDash Q$, I extend
the definition of unfounded sets with additional query atoms to
the external type systems (knowledge bases).

\begin{definition} (\textbf{Unfounded Set}) Let $P$ be a typed LP (combined KB).
Let $I$ be a partial interpretation. Let $\alpha \subseteq B_{P}$
be a set of ground atoms. $\alpha$ is an unfounded set of $P$ wrt
$I$, if for every atom $A \in \alpha$ and every ground rule
instance $A \leftarrow \beta \in ground(P)$ with a finite sequence
of ground standard literals and query atoms (querying the type
system) in $\beta$ at least one of the following conditions holds:

\begin{enumerate}
    \item at least one standard body literal $L \in \beta$ is false in $I$.
    \item at least one standard positive body literal $B \in \beta$ is contained in $\alpha$.
    \item at least one query atom $q \in \beta$ is false in $I \cup \neg \alpha$.
    \item at least one negative query atom $\sim q \in \beta$ is true in $I$
\end{enumerate}

\end{definition}

\begin{remark} Although, in this paper I do not consider Java type systems it should be noted, that the structures in Java type systems are usually not considered as interpretations in the strict model-theoretic
definition, but are composite structures involving several
different structures whose elements have a certain inner
composition. However, transformations of composite structures into
their flat model theoretic presentations is in the majority of
cases possible. However, from a practical point of view, it is
convenient to neglect the inner composition of the elements of the
universe of a structure. These elements are just considered as
"abstract" points devoid of any inherent meaning. Note that this
does not hold for procedural functions and predicates
(boolean-valued attachments / built-ins) defined on them.
\end{remark}

This structural mapping between objects from their interpretations
in the Java universe to their interpretation in the rule system
ignoring finer-grained differences that might arise from the
respective definitions is given by the following isomorphism.

\begin{definition} (Isomorphism) Let $\Sigma$ be a signature with
sorts $S_{1},..S_{n}$ and let $M_{1}$, $M_{2}$ be two
interpretations of $\Sigma$, then $f: | M_{1} | \rightarrow |
M_{2} |$ is an isomorphism of $M_{1}$ and $M_{2}$ if $f$ is a
one-to-one mapping from the universe of $M_{1}$ onto the universe
of $M_{2}$ such that:

\begin{enumerate}
    \item For every sort $S_{i}$, $m \in S_{i}^{M_{1}}$ iff $f(m)
    \in S_{i}^{M_{2}}$
    \item For every constant $c$, $f(c^{M_{1}})=c^{M_{2}}$
    \item For every n-ary predicate symbol $p$ wit n-tuple
    $m_{1},..,m_{n} \in | M_{1} |$, $\langle m_{1},..,m_{n}
    \rangle \in p^{M_{1}}$ iff $\langle f(m_{1}),..,f(m_{n})
    \rangle \in p^{M_{2}}$
    \item For every n-place function symbol $F$,
    $f(F^{M_{1}}(m_{1},..,m_{n}))=F^{M_{2}}(f(m_{1}),..,f(m_{n}))$
\end{enumerate}
\end{definition}

For instance, in ContractLog an isomorphism between Boolean Java
objects and their model-theoretic truth value is defined, which
makes it possible to treat boolean-valued procedural attachments
as body atoms in rules and establish an entailment relation as
defined above between the Java type system and the rule component.
Another example are String objects which are treated as standard
constants in rules. Primitive datatype values from the ontology
(XML) domain can be mapped similarly.

\subsection{Operational Semantics: Hybrid Polymorphic Order-Sorted
Unification}

In the following I define the operational semantics of typed
ContractLog LPs. In contrast to other hybrid approaches which
apply additional constraint atoms as type guards in the rule body
and leave the usual machinery of resolution and unification
unchanged, the operational semantics for prescriptive types in
ContractLog's typed logic is given by an order-sorted unification.
Here the specific computations that are performed in this typed
language are intimately related to the types attached to the
atomic term symbols. The order-sorted unification yields the term
of the two sorts (types) in the given sort hierarchy. This ensures
that type checks apply directly during typed unification of terms
at runtime enabling ad-hoc polymorphism of variables leading e.g.
to different optimized rule variants and early constrained search
trees. Thus, the order-sorted mechanism provides higher level of
abstraction, providing more compact and complete solutions and
avoiding possibly expensive backtracking.

The standard untyped unification algorithm in logic programming
serves as a tool for the resolution principle. It takes a set of
expressions as its input and yields the most general unifier (mgu)
for this set of formulas. A substitution $\sigma$ is called a
unifier for the set of formulas $\{E_{1},...,E_{n}$\} if
$E_{1}\sigma=E_{2}\sigma=...=E_{n}\sigma$. A unifier $\sigma$ for
the set of expressions \{$E_{1},...,E_{n}$\} is called the most
general unifier if, for any other unifier for the same set of
formulas $\theta$, there is yet another unifier $\iota$ such that
$\theta=\sigma * \iota$. For a survey on unification theory see,
e.g. \cite{Baader2001,Siekmann1989}. In the following I first
define the rules for untyped unification in terms of
equation-solving transformations \cite{Martelli1982} for
elimination (E), decomposition (D), variable binding (B) and
orientation (O). The judgements beneath the horizontal line is the
conclusion of the rule and the judgements below the line are the
premises of the rule. The computation starts with a set of
equations $E = {t_{1} = t'_{1},..,t_{n} = t'_{n}}$ representing
the terms to be unified, and to transform $E$ into the solved set
of equations $E'$ using the four given rules $E, D, B$ and $O$.
\\
(E) $\frac{E \& Y \doteq Y}{E}$, where $Y$ is a variable\\
(D) $\frac{E \& f(t_{1},..,t_{n}) \doteq f(t^{'}_{1},..,t^{'}_{n})}{E \& t_{1} \doteq t^{'}_{1} \& ...  \&  t_{n} \doteq t_{n}^{'}} $\\
(B) $\frac{E \& Y \doteq t}{\sigma (E) \& Y \doteq t}$, where $Y$
is a variable, $t$ is a constant or variable term, and $Y$ occurs
in $E$ but not in $t$, and where $\sigma = \{Y/t\}$\\
(O) $\frac{E \& t \doteq Y}{E \& Y \doteq t}$, where Y is a
variable and t is not a variable\\
\\
The computation starts with a set of equations $E = \{t_{1} \doteq
t^{'}_{1},...,t_{n} \doteq t_{n}^{'}\}$ where
$\{t_{i}/t_{i}^{'}\}$ describes the pairs of unifiable terms.
Using the four rules  $E$ is transformed into a set of equations
$E^{'} = \{Y_{i} | i \in \{1,..,n\}\}$ where $Y_{i}$ are distinct
variables which do not occur elsewhere in $E^{'}$, i.e. $\sigma =
\{Y_{1}/t_{1},..,Y_{n}/t_{n}\}$ is the most general unifier of the
unification problem given by the original set of equations $E$.
Unification fails, if there is an equation $f(t_{1},..,t_{n})
\doteq g(t_{1}^{'},..,t_{m}^{'})$ in $E$ with $f \neq g$ or if
there is an equation $Y \doteq t$ in $E$ such that $Y \doteq t$
and $Y$ occurs in $t$. I now extend this basic set of unification
rules to a hybrid polymorphic order-sorted DL-typed unification. I
restrict type checking to finding the lower bound of two types
$(r_{1},r_{2})$ under the partial order $\leq$ of the order-sorted
model with an upper bound $\top$ and a lower bound $\bot \equiv
empty$ and replace the type of a term with the more specific type
concept. Therefore, I define a \textit{lower} operation
by:\\
\\
$lower(r_{1},r_{2}) = (r_{2} / r_{1}) \rightarrow r_{1}$, if $r_{1} \leq r_{2}$ resp. $lower(r_{1},r_{2}) = (r_{1} / r_{2}) \rightarrow r_{2}$, if $r_{1} > r_{2}$\\
$lower(r_{1},\top) = (\top / r_{1})  \rightarrow r_{1}$ resp. $lower(\top,r_{2}) = (\top / r_{2}) \rightarrow r_{2}$, where $\top=untyped$\\
$lower(r_{1},r_{2}) = \bot$, otherwise, where $\bot =$ empty type.\\
\\
Note that, the operation $lower$ requires at most two queries to
the external type system to compute the lower bound of two types
having a lower bound at all. To enable polymorphic typing of
variables during typed unification, i.e. a variable may change its
type dynamically, I introduce a set
$P=\{t_{1}:r_{1},..,t_{n}:r_{n}\}$ of type restrictions, denoting
that the term $t_{i}$ (currently) has type $r_{i}$, as a prefix to
the set of equations $E$: $P \& E$. The modified and extended type rules for order-sorted unification are as follows:\\
\\
(E) $\frac{P \& E \& Y \doteq Y}{P \& E}$, where $Y$ is a variable\\
(D) $\frac{P \& E \& f(t_{1},..,t_{n}) \doteq f(t^{'}_{1},..,t^{'}_{n})}{P \& E \& t_{1} \doteq t^{'}_{1} \& ...  \&  t_{n} \doteq t_{n}^{'}} $\\
(B') $\frac{P \& Y:r \& E \& Y \doteq t}{P' \& \sigma (E) \& Y
\doteq t}$, where $Y$ is a variable, $t$ is a variable or
non-variable term, and $Y$ occurs in $E$ but not in $t$, and where
$\sigma =
\{Y/t\}$. $P \& t:r$ reduces to $P'$ using the auxiliary type rules ET and BT\\
(O) $\frac{P \& E \& t \doteq Y}{P \& E \& Y \doteq t}$, where Y
is a variable and t is not a variable\\
\\
The auxiliary rules for polymorphic unification of types are:\\
(ET) $\frac{P \& f(t_{1},..,t_{n}):r}{P}$, if $f:r_{1} .. r_{n}
\rightarrow r_{2}$ and $r_{2} \leq r$ and (ET') $\frac{P \& f(t_{1},..,t_{n}):\top}{P}$\\
(BT) $\frac{P \& Y:r_{1} \& Z:r_{2}}{P \& Y:lower(r_{1},r_{2})}$\\
\\
DL-typed unification fails \\
(1) if there is an equation $f(t_{1},..,t_{n}) \doteq
g(t_{1}^{'},..,t_{m}^{'})$ in $E$ with $f \neq g$ or \\
(2) if there is an equation $Y \doteq t$ in $E$ such that $Y
\doteq t$ and $Y \in t$ or \\
(3) if there is an equation $Y \doteq t$ in $E$ such that
$Y:r_{1}$ and $t:r_{2}$, where $t$ is a constant term and
$r_{2}>r_{1}$ or \\
(4) if there is an equation $Y \doteq Z$ in $E$ such that
$Y:r_{1}$ and $Z:r_{2}$ and $lower(r_{1},r_{2})=\perp$,
where $Y$ and $Z$ are variable terms.\\
Otherwise, if $E^{'} = \{Y_{i} | i \in \{1,..,n\}\}$ then $\sigma
= \{Y_{1}/t_{1}, .., Y_{n}/t_{n}\}$ is the mgu of the unification
problem given by the original set of equations $E$.\\
\\
In contrast to the unsorted unification, ($B'$) now involves
ad-hoc polymorphic unification of order-sorted types with a
subtype resp. equivalence test $r_{2} \leq r_{1}$ and a
computation of the lower bound of two types $lower(r_{1},r_{2})$
in the auxiliary rules, possibly assigning the more specific type
(i.e. the lower type) to a variable. The variables may change
their type during unification according to the rule ($BT$) and the
$lower$ operation. ($ET'$) is introduced to reduce unification to
special cases of the binding rule ($B$) in the untyped case
without type checking, i.e. to efficiently process untyped
variables. That is the order-sorted unification coincides with the
untyped unification, if the underlying LP does not contain typed
terms. I require that there are no infinite function definitions
such as $f(f(f(...)))$ and hence introduce the following
restriction for typed unification: $\sigma = \{X/f(t1,..,tn)\}$ if
$X \ni f$, i.e. the variable $X$ is not allowed to occur in the
function $f$ with which it is unified. Furthermore, I restrict the
unification algorithm to only well-typed terms, i.e. the type of
the argument $t_{i}$ in $f(t_{1},..,t_{n})$ must be a subtype of
the type $r_{i}$ for $f:r_{1}..r_{n} \rightarrow r$, where $r$ is
the target type of the function. I define the type of predicates,
terms and lists to be untyped by default denoted by $\top$. As a
result untyped complex terms or lists can be unified only with
other untyped variables. Informally the polymorphic order-sorted
unification rules state:

\begin{itemize}
    \item \underline{Untyped Unification:} Ordinary untyped
    unification without type checking
    \item  \underline{Untyped-Typed Unification:}The untyped query variable assumes the type of the
    typed target \item \underline{Variable-Variable Unification:}\\
    (1) If the query variable is of the same type as the target
    variable or belongs to a subtype of the target variable, the query
    variable retains its type (according to $lower$), i.e. the target
    variable is replaced by the query variable. \\
    (2) If the query variable belongs to a super-type of the target
    variable, the query variable assumes the type of the target
    variable (according to $lower$), i.e. the query variable is
    replaced by the target variable. \\
    (3) If the query and the target variable are not assignable
    ($lower=\bot$) the unification fails
    \item \underline{Variable-Constant Term Unification:} \\
    (1) If a variable is unified with a constant of its super-type,
    the unification fails. \\
    (2) If the type of the constant is the same or a sub-type of the
    variable, it succeeds and the variable becomes instantiated.
    \item \underline{Constant-Constant Term Unification:} Both constants are
    equal and the type of the query constant is
    equal to the type of the target constant.\\
    \\
    Complex terms such as lists are untyped by default and hence are
    only allowed to be unified with untyped variables.
\end{itemize}

After having defined the general semantics for order-sorted typed
unification, I will now discuss its implementation for DL type
systems in ContractLog.

A DL resp. Semantic Web type system consists of a T-Box defining
the order-sorted types and their relations and a possible empty
A-Box defining global individuals of the defined types. The T-Box
typically has a partial order $\leq$. ContractLog assumes
$owl:Resource$ as common maximum class under the partial order of
any DL type system, i.e. $\top \equiv owl:Resource \equiv
java.lang.Object \equiv untyped$ and $owl:Nothing$ to be the
minimum lower bound $\bot \equiv owl:Nothing \equiv empty$. Note
that both type system, Java and DL, coincide in the untyped
framework which can be downcasted from $java.lang.Object$ or
$owl:Resource$. The DL type checks, applied as hybrid queries to
the DL ontology type system(s) during the unification process, are
primarily concerned with \textit{Instantiation}, i.e. querying
whether an individual is an instance of a class or deriving all
individuals for a particular class, and \textit{Subsumption}, i.e.
deciding wether a class is subsumed by another class (is a
subclass of). \textit{Equivalence} inferences which check if two
classes or individuals are equal (or disjoint) and hence can be
unified is another important task which is provided by expressive
ontology languages, e.g. OWL-DL (SHOIN(D)). The typed unification
rules take into account, that the type system is defined by one or
more DL ontologies and that subtype tests via subsumption are
constrained by the expressiveness of the DL query language. For
instance, in OWL there is no way to express the concept of a most
general superclass or a most specific type. Although, it is
possible to compute such statements by applying iterative
subsumption queries, such an approach will impose greater
computational burdens for dynamic type checking. Therefore, in
ContractLog I restrict type checking to finding the lower bound of
two types $(r_{1},r_{2})$ under $\leq$ in the $lower$ operation of
the typed unification and replace the type of a term with the more
specific type concept in the unification rules. The operation
$lower$ requires at most two subsumption queries to the external
DL reasoner to compute the lower bound under the partial order
$\leq$. If the type system consists of more than one DL ontology,
the ontologies are merged into one combined ontology. The common
super class under which all ontologies are subsumed is the concept
"Resource". Hence, the partial order $\leq$ still holds for the
combined ontology under the assumption that no cycles are
introduced. Cross links between the component ontologies might be
defined, e.g. via relating classes with $owl:equivalentClass$ or
$owl:disjointWith$. Note that, this may introduce conflicts
between terminological definitions which need conflict resolution
strategies, e.g. defined by defeasible reasoning.

\begin{remark}
It is very important to note the difference between my
order-sorted typing approach and hybrid approaches which apply
additional DL atoms resp. DL constraints to query the DL ontology
component such as e.g. dl-programs \cite{Eiter2004} or Carin
\cite{Levy1996}. In my prescriptive approach the type checks in
terms of DL-queries to the DL component apply during the typed
unification process and constrain the unification of terms. The
operational semantics provides a "built-in" technical separation
between the rule inferences and the DL inferences which directly
applies during typed term unification and results in flexible
formalisms that are robustly decidable, even in the case where the
rule language is far more expressive than Datalog. This particular
combination cannot be seen neither as a super or subset of
homogeneous approaches such as SWRL nor as related to existing
hybrid approaches which apply DL constraints, since the semantics
is completely different.
\end{remark}

\section{Implementation and Integrating of Types into Rule Markup Languages}

I have implemented and evaluated my hybrid DL-typed approach (by
means of industrial use cases) in the ContractLog KR. ContractLog
is an expressive KR framework developed in the RBSLA project
(http://ibis.in.tum.de/staff/paschke/rbsla/index.htm) which
focuses on sophisticated knowledge representation concepts for
service level management (SLM) of IT services. ContractLog is
based on Prova / Mandarax, an open-source Java-based
backward-reasoning rule engine with a Prolog like scripting syntax
(http://www.prova.ws/). The ContractLog KR implements several
logical formalisms such as event logics, defeasible logic, deontic
logics and comes with a novel sequential resolution for extended
LPs with extended defeasible Well-founded Semantics. The rule
engine (an extension to the Prova inference engine) coming with
ContractLog/RBSLA distribution
(https://rbslasourceforge.net/projects/rbsla) performs query
answering on the hybrid rules. As an external service I use the
Jena API (http://jena.sourceforge.net) in combination Pellet
(http://www.mindswap.org/2003/pellet/), integrated as an external
DL reasoner. During resolution non-grounded facts with variables
are interpreted as instance queries on the external service to
decide wether an individual is an instance of a class given by the
type of the bound variable or to derive all individuals explicitly
introduced in the DL A-Box which are instances of the type (class)
of the free variable, which becomes instantiated with the found
individuals. During unification dynamic type checking must be
performed via subsumption queries using the external DL reasoner.
To achieve decidability I restricted the rule component to LPs
with finite functions and the DL component to decidable DLs
reaching from \textit{ALC} to \textit{SHIQ(D)} resp.
\textit{SHOIN(D)}. The only restriction is on the interchange of
entailments between both formalisms in the sense that there is a
unidirectional flow from the DL part to the LP part. In particular
this means, that only consequences involving individuals which are
explicitly introduced in the A-Box are passed from the DL reasoner
to the LP reasoner, but new individuals can be introduced within
the LP facts or rule (heads) which do not affect query answering
with the DL reasoner. The complexity for reasoning in my hybrid
DL-typed logic is derived from the complexity of the component
languages and the applied semantics for DL languages and extended
LPs under WFS. The worst case complexity of reasoning in my hybrid
DL-typed logic is \textit{EXPTIME} for a combination of extended
LPs and OWL Lite and \textit{NEXPTIME} for a combination with
OWL-DL. Note, that Jena also support RDFS reasoning and I also
allow to define type systems in RDFS. Since, the RDFS reasoner
omits many complex
entailments such as equality reasoning , using RDFS to define light-weight type systems will lead to more efficient type unification.\\
\\
Theorem 1: \textit{Query answering in hybrid DL-typed Datalog is
EXPTIME-complete for OWL-Lite (SHIQ(D)) type systems resp.
NEXPTIME-complete for OWL-DL (SHOIN(D)) type systems.}\\
\\
The ContractLog KR based on Prova provides a Prolog related
scripting syntax to write LPs, e.g.

\begin{verbatim}
import("./dl_typing/WineProjectOWL.owl").
reasoner("dl"). % OWL DL reasoner

serve(X:vin_Wine):- wine(X:vin_Wine).
wine(vin_White_Wine:Chardonnay). wine(X:vin_Red_Wine).
\end{verbatim}

The example defines a rule, where the variable $X$ is of type
$vin\_Wine$ ($vin$ is the namespace prefix). The first fact
introduces a new individual (constant term) $Chardonny$ which is
of type $vin\_White\_Wine$ and hence unifies with $X:vin\_Wine$.
The second fact defines a variable $X$ of type $vin\_Red\_Wine$
which is used to query all individuals of type $vin\_Red\_Wine$
from the DL A-box using the external DL reasoner, if $X$ is a free
variable resp. to check whether $X$ is of type $vin\_Red\_Wine$,
if $X$ is a previously bound variable. The type system is imported
via the $import(<URI>)$ predicate. Different external reasoner
such as $"transitive"$ (simple transitive reasoner), $"rdfs"$
(rdfs reasoner), $"owl"$ (owl lite reasoner), $"dl"$ (owl dl
reasoner) etc. are supported within the ContractLog implementation
(see OWL2PROVA framework in the RBSLA/ContractLog distribution
available at https://rbslasourceforge.net/projects/rbsla).

In addition to this Prolog related scripting syntax I provide a
XML/RDF serialization based on RuleML (http://www.ruleml.org/).
RuleML (Rule Markup Language) is a standardization initiative with
the goal of creating an open, producer-independent XML/RDF based
web language for rules. I use the $@type$ attribute to define
typed terms.

\begin{verbatim}
<Implies>
  <head> <Atom>
    <Rel>serve</Rel>
    <Var @type="vin:WhiteWine">X</Var>
  </Atom></head>
  <body><And><Atom>
     <Rel>wine</Rel>
     <Var @type="vin:WhiteWine">X</Var>
  </Atom></And></body>
</Implies>
\end{verbatim}

The RuleML documents are transformed into the target execution
language, i.e. into ContractLog/Prova scripts and then executed in
the ContractLog KR using the typed unification to process hyrbid
DL-typed rules. I have implemented a XSLT stylesheet which
transforms typed RuleML LPs into typed ContractLog scripts.

\section{Summary}

The main motivation for introducing Semantic Web or Java based
types into declarative logic programs comes from Software
Engineering, where principles such as data abstraction,
modularization and consistency checks are vital for the
development and maintenance of large rule bases. Distributed
system engineering and collaboration, where domain-independent
rules need to be interchanged and given a domain-dependent meaning
based one or more common vocabularies in their target environments
is an other. The possibility to use arbitrary Semantic Web
ontologies and object-oriented domain models in declarative logic
programming gives a highly flexible, extensible and syntactically
rich language design with a precise semantics. Semantic Web
counterparts of domain-specific ontologies, e.g. OWL Time, can be
easily used as type systems giving the rule terms a
domain-specific meaning. This syntactic and semantic combination
which allows efficient declarative programming is vital for modern
Semantic Web based rule systems. From a computational point of
view, the use of order-sorted types can drastically reduce the
search space, hence increasing the runtime efficiency and the
expressive power, e.g. enabling overloading of rule variants. The
tight combination of declarative and object-oriented programming
via rich Java-based procedural attachments facilitates integration
of existing procedural functionalities, tools and external data
sources into rule executions at runtime. In this section I have
presented a hybrid approach which provides a technical separation
with minimal interfaces between the rule and type components
leading to a robust, flexible and expressive typed logic with
support for external Java and DL type systems, Java-based
procedural attachments, modes and built-ins. The implementation
follows a prescriptive typing approach and incorporates type
information directly into the names of symbols in the rule
language. The interaction between the rules and the type system is
based on entailment in a multi-sorted logic. As an operational
semantics a typed unification is applied which permits dynamic
type checking, ad-hoc polymorphism, i.e., variables might change
their types (coercion), and overloading, i.e. overloading  of
rules by using different types in their rule heads, leading to
variants of the same rule.

\bibliography{bibliography}
\bibliographystyle{alpha}
}

\end{document}